\documentclass[10pt, a4paper]{article}
\usepackage{lrec}
\usepackage{multibib}
\newcites{languageresource}{Language Resources}
\usepackage{graphicx}
\usepackage{tabularx}
\usepackage{algorithm}
\usepackage[noend]{algpseudocode}

\usepackage{titlesec}
\titleformat{\section}{\normalfont\large\bf\center}{\thesection.}{1em}{}
\titleformat{\subsection}{\normalfont\SmallTitleFont\bf\raggedright}{\thesubsection.}{1em}{}
\titleformat{\subsubsection}{\normalfont\normalsize\bf\raggedright}{\thesubsubsection.}{1em}{}
\renewcommand\thesection{\arabic{section}}
\renewcommand\thesubsection{\thesection.\arabic{subsection}}
\renewcommand\thesubsubsection{\thesubsection.\arabic{subsubsection}}

\usepackage[utf8]{inputenc}

\usepackage{hyperref}
\usepackage{xstring}

\usepackage{color}

\usepackage{booktabs} %
\usepackage{url}
\usepackage{microtype}
\usepackage{amsmath}

\usepackage{etoolbox}

\newcommand{\includeifspace}[1]{#1}

\newcommand{\freq}{\mathrm{f}}

\title{Correcting the Autocorrect: Context-Aware\\
Typographical Error Correction via Training Data Augmentation}

\name{Kshitij Shah, Gerard de Melo}

\address{Department of Computer Science\\
         Rutgers University--New Brunswick\\
         kshitij.shah@rutgers.edu, gdm@demelo.org}

\abstract{
In this paper, we explore the artificial generation of typographical errors based on real-world statistics. We first draw on a small set of annotated data to compute spelling error statistics. These are then invoked to introduce errors into substantially larger corpora.
The generation methodology allows us to generate particularly challenging errors that require context-aware error detection.
We use it to create a set of English language error detection and correction datasets.
Finally, we examine the effectiveness of machine learning models for detecting and correcting errors based on this data. The datasets are available at \url{http://typo.nlproc.org}.
\\ \newline \Keywords{Corpus, Error Generation, Deep Learning } }

\begin{document}

\maketitleabstract

\section{Introduction}

\includeifspace{With the emergence of the Internet as the preferred medium for communication and the rise of social media,
the number of words typed by an average human being has increased manyfold in the last decades.}
The majority of this communication is casual in nature, with less attention and time dedicated by the writer compared to formal writing. 
This leads to a plethora of typographical errors in such text. 
This problem is exacerbated by the comparably small size of the keyboard on hand-held devices.
Hence, typographical errors, which were somewhat more limited and likely to be corrected immediately in past decades, are now ubiquitous. 

Traditional solutions such as dictionary-based spelling correction are often inadequate. First of all, online writing tends to make liberal use of
neologisms, slang words, and so on, that are lacking in pre-existing dictionaries. Any automated means to mine dictionaries from social media risk also incorporating frequent misspellings such as *\emph{seperate} into the dictionary. Second, a higher error rate also entails a larger likelihood of a typographical error leading to a form that happens to be a legitimate word in the dictionary. This problem exists even with more advanced spelling correction software, as indeed, many such errors are now \emph{caused} by autocorrection software.

\includeifspace{To overcome these challenges, 
an alternative is to draw on advances in machine learning to identify and correct such typographical errors.}
While autocorrection typically considers just the current and possibly the last few words, deep models that can account for both sides of a larger context window have the potential to more accurately determine whether a word fits in context. Unfortunately, deep learning generally requires large annotated corpora for effective training. Such large annotated datasets are difficult to procure, as the labeling process tends to be time-consuming and expensive \cite{QiuXuZhangWangShenDeMeloLongLi2020DataAugmentation}. 
\includeifspace{There have been significant efforts in the last decade to overcome this gap for the task of grammatical error correction \cite{dale2011helping,ng2013conll,ng2014conll}. However, no sufficiently large datasets exist for typographical error correction.}

In this paper, we aim to generate realistic typographical errors based on statistical distributions collected from a relatively small annotated seed dataset. Our method captures the error patterns from the seed data and generates similar error distributions on error-free target corpora of arbitrary size.
An important special case, particularly when typing on hand-held devices, is the abundance of \emph{real-word errors}, also known as \emph{atomic typos}. 
These occur when misspelled words happen to also be valid words in the dictionary, but used in the wrong context, requiring
\emph{context-aware spelling correction}.
We examine inducing such errors automatically via a dictionary-based spelling corrector. 
Finally, we explore the effectiveness of deep neural networks to detect and correct such errors.
While the experiments in this paper are limited to English, our method is applicable to any language with a restricted (true) alphabet.

\section{Related Work}

Due to the high cost and difficulty of labeling large text datasets for error correction, several studies explored generating artificial grammatical mistakes
\cite{foster2009generrate,felice2014generating,ng2013conll,rei2017artificial,kasewa-etal-2018-wronging}.
\includeifspace{For example, \newcite{foster2009generrate} achieved this by moving, substituting, inserting, and removing words in a sentence, and investigated part-of-speech tags to induce more realistic errors. \newcite{felice2014generating} used probability word-level statistics computed on the CoNLL 2013 shared task data \cite{ng2013conll} to introduce artificial grammatical errors at the word level. \newcite{rei2017artificial} used a statistical machine translation system to translate correct text to ungrammatical text. 
For training, they relied on the annotated dataset from CoNLL 2014 \cite{ng2014conll} but considered in inverted order: They feed the corrected version as input and the ungrammatical one as the output.}
These works aim to mimic the phenomenon of grammatical errors, i.e., when linguistic utterances are deemed lacking in grammatical acceptability with respect to a linguistic standard. 

Only very little research has been conducted on simulating errors originating from basic typographical mistakes made by a writer. While these may stem from a lack of knowledge of the authoritative spelling or appear as a symptom of conditions such as dyslexia, they may also be caused by simple key entry mistakes such as those referred to as \emph{fat finger errors}.

\includeifspace{In the past, such errors were often corrected by means of a dictionary-based spelling corrector, but with the proliferation of hand-held devices, this no longer seems sufficient. Indeed, many such devices invoke autocorrection tools, which may lead to entirely new errors that are very hard to detect.}
\newcite{bigert2003autoeval} developed a tool called `Missplel', which could introduce character-level errors, among others. However, those errors were not modeled upon real-world data. \newcite{whitelaw2009using} repurposed Web data as a noisy corpus for spelling correction and used a limited corpus with artificially inserted misspellings to tune classifiers. \newcite{ghosh2017neural} created a synthetic dataset of incorrect key strokes by sampling from a Gaussian at each key location on a virtual keyboard. They also created another dataset by replacing correct words with their misspellings, as given by an annotated typo corpus \cite{typocorpus}. Their paper noted the lack of datasets for this problem, which led them to create their own to train and test their model.
Our work differs from their approach in that we induce a noise model from text that can then be used to introduce a wide range of errors, instead of replacing a small set of words by their misspellings. \newcite{baba2012spelling} studied different error categories arising during typing.

Methods for context-aware spelling correction have mostly focused on small sets of words typically confused in human writing: Many of the most well-known approaches to this task rely on predefined \emph{confusion sets} to solve it \cite{carlson2007memory,golding1999winnow,carlson2001scaling,banko2001scaling}. However, with the prominence of autocorrect systems on mobile devices, this approach is ineffective. Hence, we explore generating realistic spelling errors requiring context-aware correction in this paper.

\section{Method}
\label{sec:method}

Our method for artificial error generation consists of two phases. First, we compute probabilities for different types of errors based on a small annotated corpus. Subsequently, we rely on these statistics to generate errors in the target text.  

\subsection{Model Induction}
We classify typographical errors into four categories and induce a model consisting of a series of probability distributions. Errors of each of these types are counted in an annotated corpus and a specific set of rules are used to calculate probabilities for each category. In many cases, probability distributions are computed by counting a specific error pattern with regard to different characters, and then normalizing by dividing counts by the total number of errors of that category.

\paragraph{Substitution Errors.} A substitution error occurs when another character replaces the correct character. Intuitively, a character is more likely to be replaced by a nearby character on a QWERTY keyboard. Moreover, some characters are more likely to be mistyped compared to others. We empirically determine the probability of each character $c$ being substituted $P(\mathrm{substitution} \mid c)$, as well as the probability of each character $c'$ in the character set $C$ replacing $c$, i.e., $P_{\mathrm{substitution}}(c' \mid c)$.
\[
P(\mathrm{substitution} \mid c) = \frac{\freq_{\mathrm{substitution}}(c)}{\freq(c)}  
\]

\[
P_{\mathrm{substitution}}(c' \mid c) = \frac{\freq_\mathrm{substitution}(c', c)}{\displaystyle\sum_{\overline{c} \in C} \freq_\mathrm{substitution}(\overline{c}, c)}  
\]

where $\freq_{\mathrm{substitution}}(x)$ denotes the frequency of character $x$ being replaced, while $\freq_{\mathrm{substitution}}(y, x)$ denotes the frequency of character $y$ replacing character $x$.

\paragraph{Insertion Errors.} Insertion errors occur when an additional character is inserted by mistake. We assess the probability of a given inserted character depending on the adjacent characters, assuming that these typically result from nearby keys being activated near-simultaneously. However, the added character could be attributed to either the previous or to the next character. We resolve this by attributing it to the adjacent character that is nearer to it on a virtual keyboard.
If the distance to both neighbors on the virtual keyboard is the same, we assign it to one of them randomly. 
The computed statistics include probabilities of any character being inserted before or after a given character $c$, i.e., $P(\mathrm{insertion} \mid c)$, and the individual probabilities for the inserted character, given by $P_{\mathrm{insertion}}(c' \mid c)$. The probabilities for insertion before and after a given character are computed separately. 

\[
P(\mathrm{insertion} \mid c) = \frac{\freq_\mathrm{insertion}(c)}{\freq(c)}  
\]

\[
P_\mathrm{insertion}(c' \mid c) = \frac{\freq_\mathrm{insertion}(c', c)}{\displaystyle\sum_{\overline{c} \in C} \freq_\mathrm{insertion}(\overline{c}, c)}  
\]

where $\freq_{\mathrm{insertion}}(x)$ is the frequency of any character being inserted adjacent to $x$, and $\freq_{\mathrm{insertion}}(y, x)$ denotes the frequency of character $y$ being inserted adjacent to character $x$.

Another caveat is when the inserted character is the same as its neighbor, i.e., the same character is typed twice. In this case, we cannot easily decide which of these two occurrences ought to be considered as the added one. To address this issue, we define the distinct subcategory of replication errors, which we measure separately.

\paragraph{Replication Errors.} A replication error occurs when a character is repeated twice. We compute a separate probability of replication $P(\mathrm{replication} \mid c)$ given each character. %

\[
P(\mathrm{replication} \mid c) = \frac{\freq_\mathrm{replication}(c)}{\freq(c)}  
\]

\paragraph{Deletion Errors.}  A deletion error occurs when a user misses a particular character that should have been entered.  Some characters exhibit a higher tendency of getting missed. Hence, we compute the probability $P(\mathrm{deletion} \mid c)$ of each character $c$ getting missed. 

\[
P(\mathrm{deletion} \mid c) = \frac{\freq_\mathrm{deletion}(c)}{\freq(c)}  
\]

\paragraph{Transposition Errors.} A transposition error is registered when two consecutive characters are typed out of order. Only consecutive character pairs are considered for this type of error and the probability of transposition for every possible sequence of two characters $c_1$ and $c_2$ is represented as $P(\mathrm{transposition} \mid c_1c_2)$.

\[
P(\mathrm{transposition} \mid c_1c_2) = \frac{\freq_\mathrm{transposition}(c_1c_2)}{\freq(c_1c_2)}
\]

\subsection{Error Generation}

\paragraph{Corpus Cleaning.} Before introducing errors, we seek to ensure that the original corpus does not have substantial typographical errors. Thus, in a preprocessing step, given a dictionary vocabulary $V$, documents with out-of-vocabulary words not in $V$ are discarded.

\paragraph{Error Induction.} Our error generation process is based on the assumption that typing proceeds progressively character by character and that errors may occur at every keystroke. Hence, the errors are generated at each character based on the precomputed statistics. Each error category is considered individually in random order with its respective probability multiplied by a weighting coefficient for that error category. 
The weighting coefficients allow us to control the relative frequency of each error category individually as well as the overall error rate. Retaining a uniform distribution of coefficients mimics the original distribution observed on annotated data.
These coefficients also enable us to the use of character frequency distributions from a corpus other than the one used to compute error statistics. This later allows us to compute probability values based on the character distribution of the target corpus we are inducing errors on, while still using the error distribution from the annotated corpus.
In the case of insertion or substitution errors, if an error is determined to be generated, the candidate for insertion or substitution is selected based on the computed statistics.
If a particular error category is applied to a character, the remaining error categories are no longer considered for it.

\paragraph{Error Replacement.} Having generated the corrupted text, we invoke a dictionary-based %
spelling correction algorithm to generate confused words that are valid words but unlikely to fit in the context. Specifically, confusion is enforced by selecting the highest-ranked suggestion that is not correct.\footnote{Another variant of the data, which picks the best suggestion without forcing confusion is also generated, but not used in the subsequent experiments presented in this paper. It will be available to the community as part of the dataset.} 
However, if Enchant makes only one suggestion for a given error, it is accepted without forcing confusion.

\includeifspace{This results in hard to detect errors calling for context-aware spelling correction.
}

\paragraph{Output Dataset.} During this entire process, the number of tokens is kept the same as in the original text. For single character words and rarely for other short words, it is possible that an entire word is removed during the character-level corruption. In such cases a placeholder token is used in its place. The placeholder is a special symbol such as  $<$UNK$>$.
If a word is split in two by the spelling corrector, the longer one is chosen and the shorter one discarded. For instance, if \emph{blike} is corrected as \emph{be like}, we only consider \emph{like} and discard \emph{be}. This constraint facilitates working with our data, because the labels are on every word and are easier
to process if the number of words does not
diverge between the actual text and our generated version. 
Thus, in the final error detection datasets that we induce, each word has a binary label associated with it. Introduced errors are labeled as positive, while original words are labeled as negative.
We also provide the original word as the ground truth for the task of error correction.

\section{Analysis of Error Statistics}
\label{sec:stats}

We derive error statistics from the Twitter Typo Corpus \cite{typocorpus}, which contains 39,171 pairs of words with typographical errors along with the correct word. While the dataset only includes lower-case English letters, our method can generalize to any character set.

\begin{figure}[!h]
\begin{center}
\includegraphics[width=0.5\textwidth]{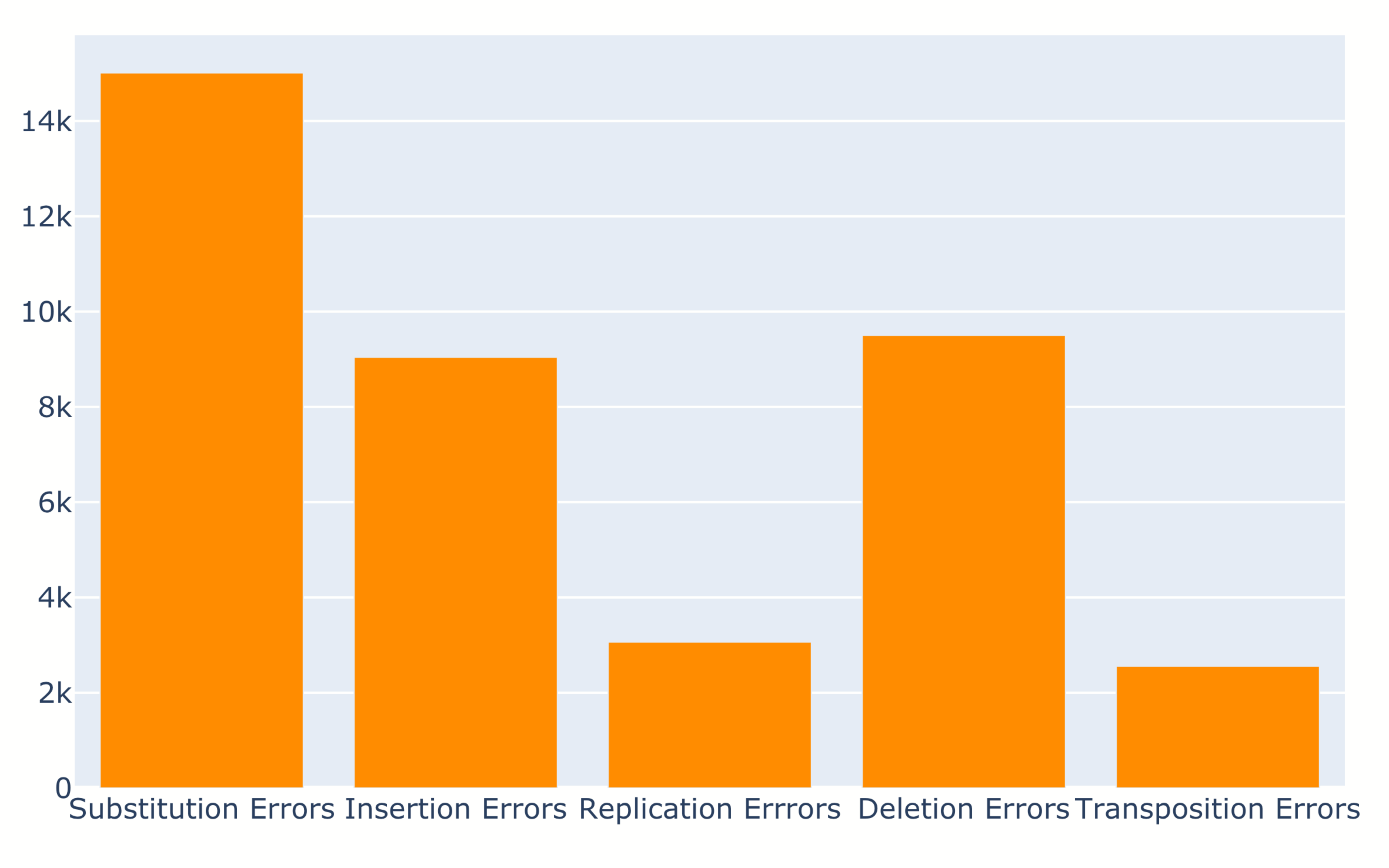} 
\caption{Distribution of Error Types}
\label{fig.dist}
\end{center}
\end{figure}

Figure \ref{fig.dist} shows the distribution of errors among the error types described in Section \ref{sec:method} in the Twitter Typo Corpus \cite{typocorpus} used for error induction. We found that the error distribution is dominated by substitution, insertion, and deletion errors, while the replication and transposition errors are relatively scarce. 
Note that the latter two can be considered derivative. For instance, replication errors can be viewed as insertion errors with the inserted character being the same as an adjacent character.

\begin{figure}[!h]
\begin{center}
\includegraphics[width=0.5\textwidth]{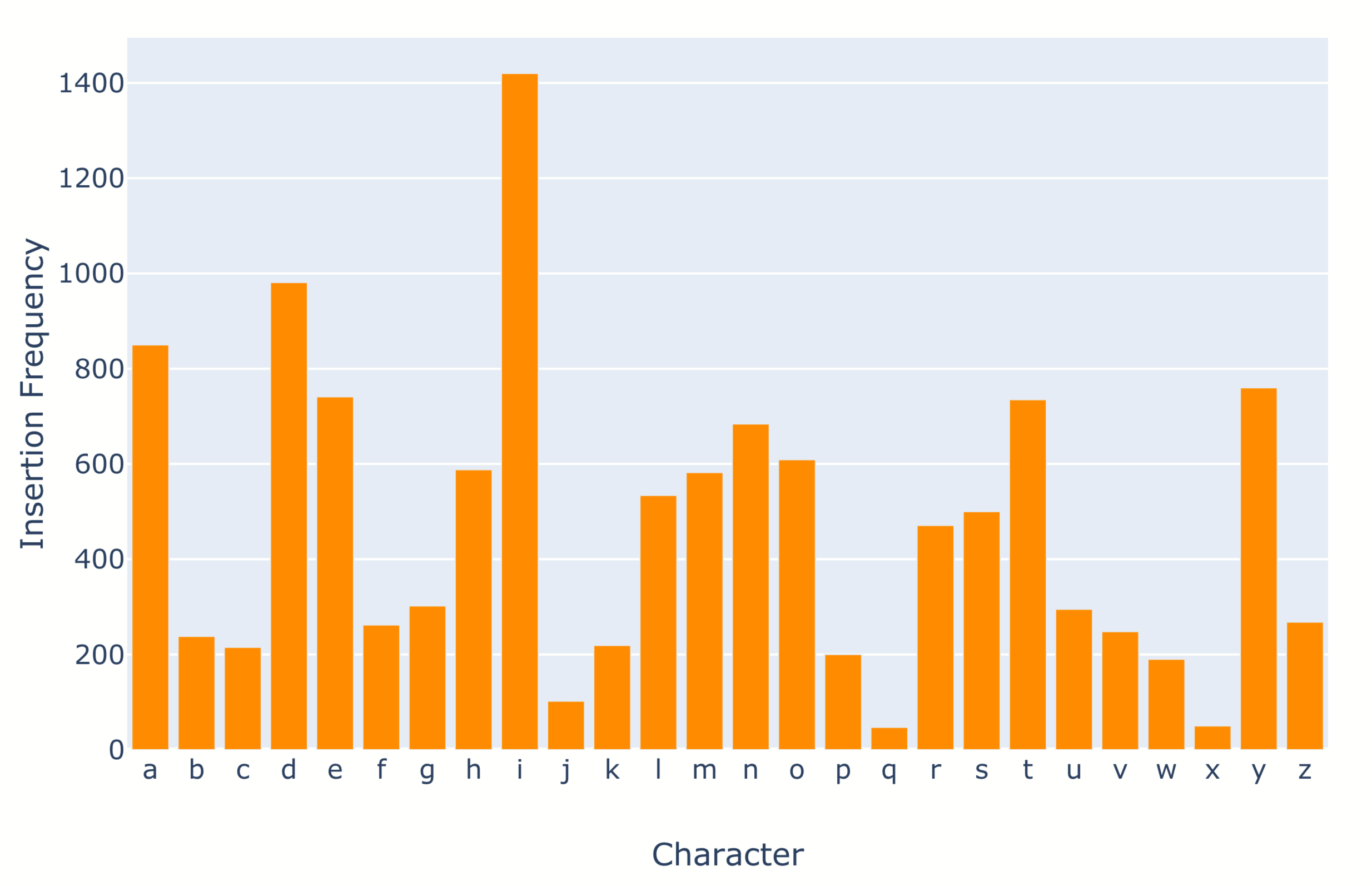} 
\caption{Frequency of Insertion Errors by Characters}
\label{fig.InsertFreq}
\end{center}
\end{figure}

\begin{figure}[!h]
\begin{center}
\includegraphics[width=0.5\textwidth]{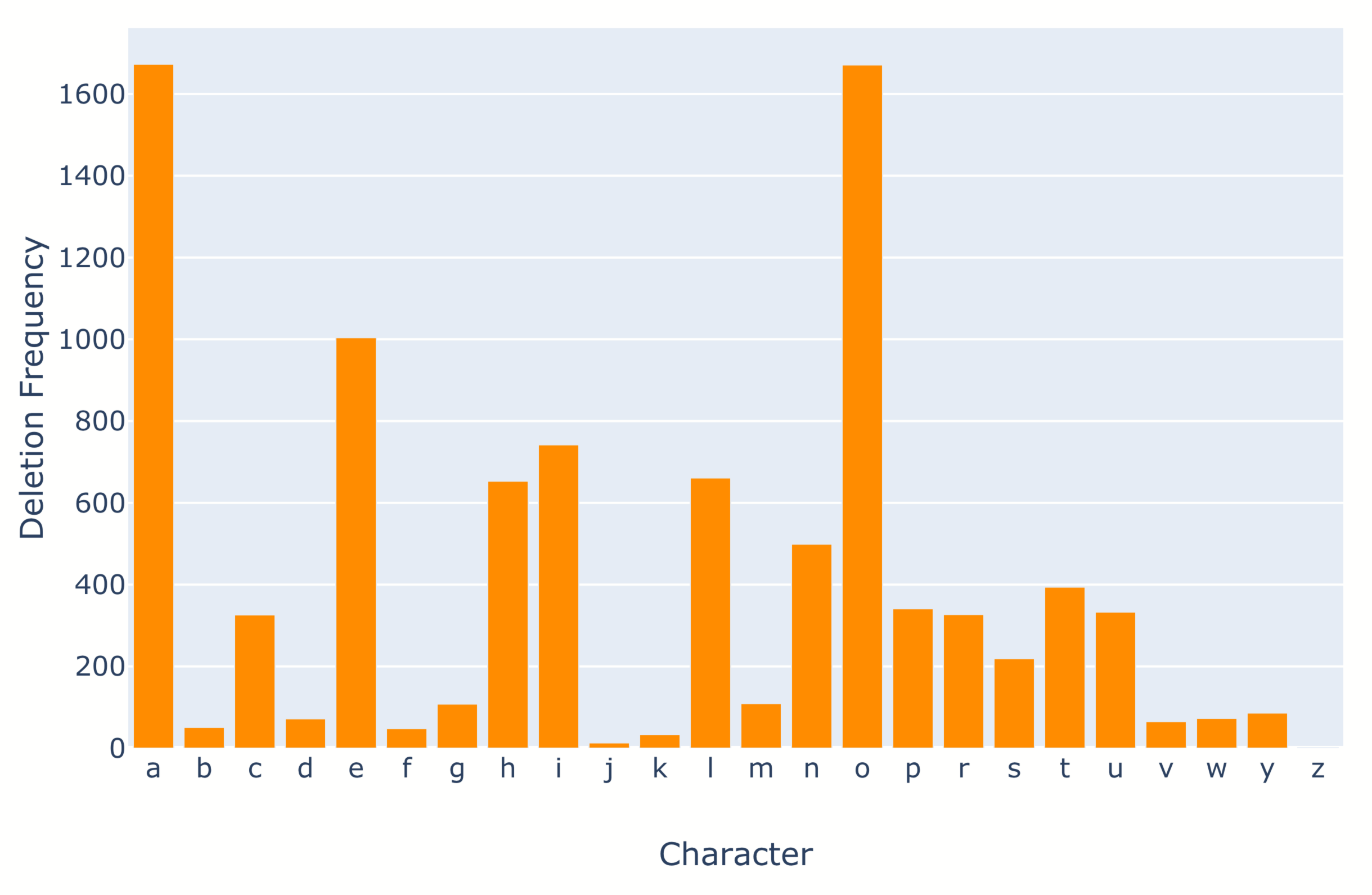} 
\caption{Frequency of Deletion Errors by Characters}
\label{fig.DeleteFreq}
\end{center}
\end{figure}

\begin{figure}[!h]
\begin{center}
\includegraphics[width=0.5\textwidth]{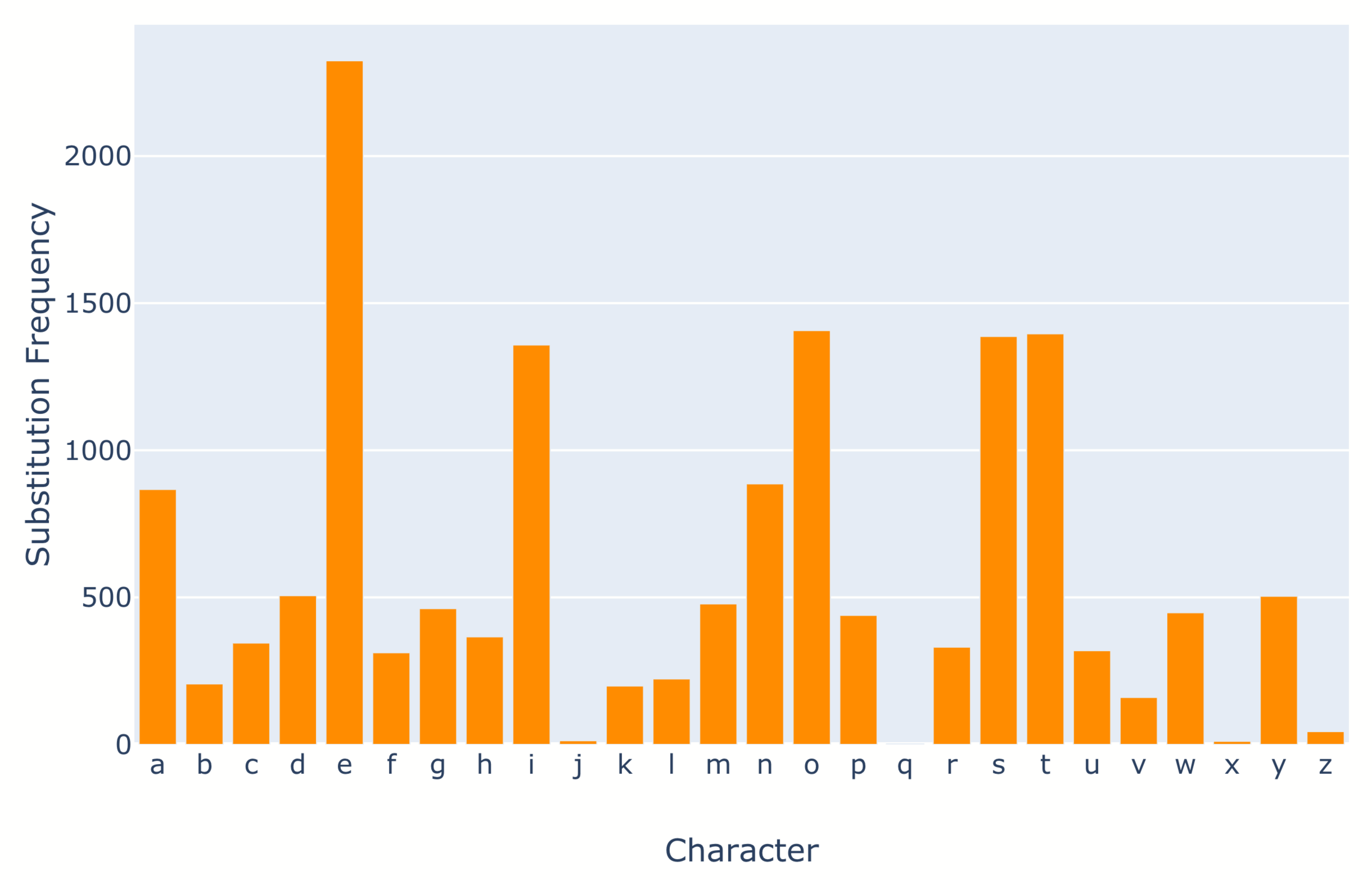} 
\caption{Frequency of Substitution Errors by Characters}
\label{fig.subFreq}
\end{center}
\end{figure}

\begin{figure}[!h]
\begin{center}
\includegraphics[width=0.5\textwidth]{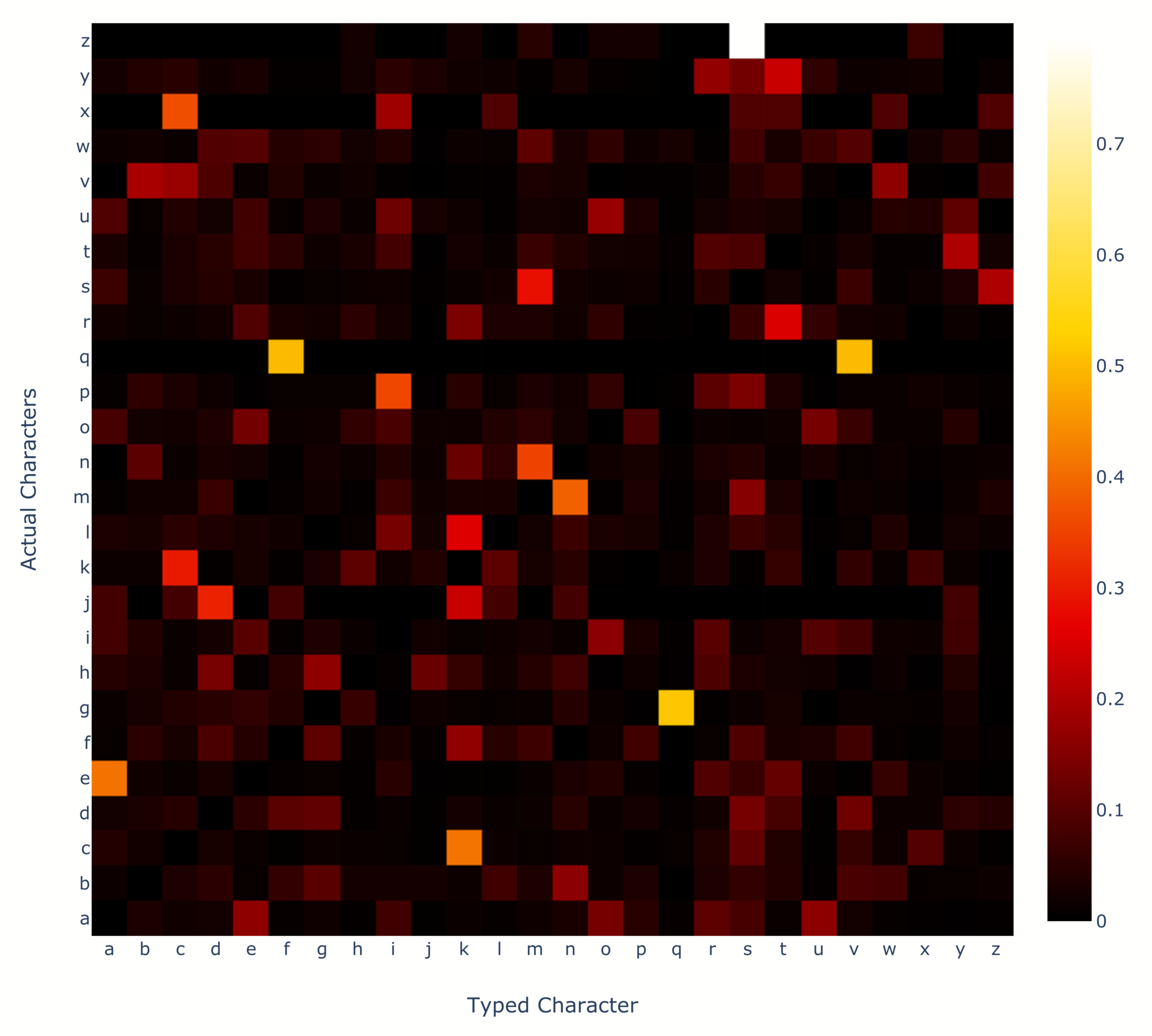} 
\caption{Substitution Probabilities between Characters}
\label{fig.subProb}
\end{center}
\end{figure}

Figure \ref{fig.InsertFreq} shows the frequency distribution of inserted characters. We observe several patterns emerging from the data. The characters that on standard QWERTY keyboards are located near the finger tips in a natural typing position are more likely to be inserted by mistake. For example, `a', `d', `e', and `i' exhibit high insertion error frequencies.
Figure \ref{fig.DeleteFreq} shows the frequencies for deletion errors, which accounts for a certain character being missed by the writer. The deletion frequency is highly correlated with the natural occurrence frequency of characters. Frequently used vowels such as `a', `e', `i', `o' show a higher deletion frequency. However, this does not imply that they are more likely to be missed. Our model combines these statistics with the occurrence frequencies and error rate hyperparameter in order to generate probabilities of deletion errors. A noticeable characteristic that emerges is that the characters in the middle row of QWERTY keyboards such as `d', `f', `g', `j', `k' are less likely to be missed.
Figure \ref{fig.subFreq} depicts the frequencies of different characters being replaced by another character, i.e., substitution errors. This type of error also exhibits correlation with the occurrence frequencies of characters. This statistic is used to determine the probability of a substitution error at a given character during error induction. Figure \ref{fig.subProb} shows the probabilities of particular characters being replaced by particular other characters, given that a substitution error is present. These probabilities are independent of the error rate and are only applied once a substitution error is determined.

\section{Experiments}

\paragraph{Input Data.}

We rely on the statistics derived from the Twitter Typo Corpus \cite{typocorpus}, described in Section \ref{sec:stats} to introduce
errors into two corpora, a food review corpus, and a large movie review one. For food reviews, we consider the Amazon fine food review dataset, which consists of 568,454 food reviews collected from Amazon, along with metadata \cite{mcauley2013amateurs}. The large movie review dataset \cite{maas2011learning} contains 50,000 labeled and 50,000 unlabeled movie reviews collected from IMDB. We only use the text from both of these datasets.

\begin{table}[tbp]
\centering
\footnotesize
\begin{tabular}{lrr}
  \toprule
       Error Level & \multicolumn{2}{c}{Corrupted Words (\%)}\\
   \cmidrule{2-3}
    (\% of corrupted char) & Amazon  &  IMDB  \\
  \midrule
  Low (3.75\%) & 15.73 & 15.75 \\
  Medium (7.5\%) & 28.66 & 27.27 \\
  High (15\%) &  48.27 & 42.07 \\
  \bottomrule
\end{tabular}
\caption{Generated Error Percentage}
\label{error-levels}
\end{table}

\paragraph{Error Corpus Induction.}
First of all, to remove pre-existing typographical errors from these corpora, we construct a dictionary $V$ from the Enchant spell checker, enhanced by the vocabulary of GloVe embeddings trained on 6 billion words from Wikipedia and Gigaword \cite{pennington2014glove}.\footnote{We do not rely on embeddings trained on CommonCrawl, as Web data contains substantially more misspelling forms.}
On the fine food review dataset, we also remove reviews that are outliers in terms of their length.\footnote{Specifically, those with a character length three standard deviations above or below mean. Hence, we filter out reviews longer than 1,715 characters, but no reviews with shorter length, as three standard deviations below mean is less than 0.}
This leads to 254,638 samples in the fine food review dataset, from which we sample 160,000 for training, 40,000 as a held-out validation set, and 50,000 as test data. The total size of the dataset %
is 1.13 GB.
From the movie review data, we only exclude reviews with out-of-vocabulary words. The dataset already provides training and test splits, and we reserve 20\% of the former for validation, leading to 42,452 training, 10,613 validation, and 17,915 test samples. Despite having fewer sentences, the average sentence length is longer, resulting in a size of 1.27 GB.

We use our proposed method to introduce three levels of error rates, independent of one another. At the highest error rate setting, we induce errors in 15\% of the total characters. The other two levels are 7.5\% and 3.75\%, respectively. For generating confused words, we selected the the highest ranked incorrect suggestions by the Enchant spell-checker \cite{enchant}. This allow us to consistently obtain challenging real-word errors.
Table \ref{error-levels} shows the percentage of words with spelling errors corresponding to each level.

For analysis purposes, we evaluate separate neural models as baselines to detect and correct these errors. 
This allows us to evaluate the detection and correction independently and use the most suitable model for each task.

\includeifspace{
\begin{table}[htbp!]
\centering
\footnotesize
\begin{tabular}{lr}
  \toprule
  Hyperparameter &  Value\\
  \midrule
  Number of Epochs & 50 \\
  Learning Rate & 0.005 \\
  Optimizer & Adam \\
  Embedding Dimensions & 100 \\
  Recurrent Units & 100 \\
  Dropout & 0.5 \\
  Mini-batch Size & 512 \\
  \bottomrule
\end{tabular}
\caption{Hyperparameters for Error Detection}
\label{hyperparam-table}
\end{table}

\begin{table}[htbp!]
\centering
\footnotesize
\begin{tabular}{lr}
\toprule
Hyperparameter &  Value\\
\midrule
Number of Steps & 5,000 \\
Number of Layers & 6 \\
Number of Units & 512 \\
Number of Heads & 8 \\
FFN Dimensions & 2,048 \\
Attention Dropout & 0.1 \\
FFN Dropout & 0.1     \\   
  \bottomrule
\end{tabular}
\caption{Hyperparameters for Error Correction}
\label{transformer-hyperparam-table}
\end{table}
}

\begin{table}[tbp]
\centering
\footnotesize
\begin{tabular}{lrrrr}
  \toprule
   Dataset (Error Level) & Recall & Precision & $F_1$ score \\
  \midrule
  Amazon (3.75\%) & 0.8867 & 0.9298 & 0.9077  \\
  Amazon (7.50\%) & 0.9105 & 0.9448 & 0.9273 \\
  Amazon (15.00\%) & 0.9329 & 0.9560 & 0.9433 \\
  IMDB (3.75\%) & 0.8172 & 0.8576 & 0.8369 \\
  IMDB (7.50\%) & 0.8673 & 0.8901 & 0.8786 \\
  IMDB (15.00\%) & 0.9105 & 0.9155 & 0.9130 \\
  \bottomrule
\end{tabular}
\caption{Effectiveness of Bidirectional LSTM for Context-Aware Error Detection}
\label{tab:performance-eval}
\end{table}

\paragraph{Spelling Error Detection.}
For context-aware spelling error detection, we evaluate the effectiveness of a Bidirectional LSTM based sequence labeling model
to predict the probability of every word's label being positive, with a sigmoid activation function. %
We initialize our embeddings with 100-dimensional GloVe vectors trained on Wikipedia and Gigaword \cite{pennington2014glove} and make it further trainable. Out-of-vocabulary words are initialized randomly with the same center and scale as the GloVe vectors. The recurrent layer contains 100 LSTM units. We also employ Dropout with a rate of 0.5 to avoid overfitting. For training, we use Adam optimization with a learning rate of 0.005 for 50 epochs on mini-batches of size 512, with early stopping enabled. To address the class imbalance and to achieve maximum effectiveness, the threshold for positive prediction is fit on the validation set to maximize the $F_1$ score. 
\includeifspace{Hyperparameters are given in Table \ref{hyperparam-table}.
} %
Table \ref{tab:performance-eval} shows the effectiveness of the detection model on the test data. We observe that with higher error rates, the $F_1$ score is also higher, in part because the $F_1$ measure is not completely resilient to class imbalance. We also observe a difference in effectiveness between the two datasets, which may stem from the smaller training size for the IMDB dataset combined with its longer sequence lengths.

\begin{table}[htbp!]
\centering
\footnotesize
\begin{tabular}{lrrr}
  \toprule
       & \multicolumn{2}{c}{BLEU Score}\\
    \cmidrule{2-3}
   Dataset (Error Level) &  Noisy & Corrected & Gain \\
  \midrule
  Amazon (3.75\%) & 0.6491 & 0.8685 & 0.2194 \\
  Amazon (7.50\%) & 0.4263 & 0.8107 & 0.3844 \\
  Amazon (15.00\%) & 0.1878 & 0.6661 & 0.4783 \\
  IMDB (3.75\%) & 0.6378 & 0.6852 & 0.0474 \\
  IMDB (7.50\%) & 0.4122 & 0.4169 & 0.0047 \\
  IMDB (15.00\%) & 0.1778 & 0.4806 & 0.3028 \\
  \bottomrule
\end{tabular}
\caption{Effectiveness of Transformer Network for Context-Aware Spelling Error Correction}
\label{tab:correction-eval}
\end{table}

\paragraph{Context-Aware Spelling Error Correction.} %
To evaluate our data on error detection as well as correction, we rely on a Transformer model \cite{vaswani2017attention} to transform the noisy sequence with errors to the correct sequence. The intuition is that the model will consider the wider context to detect and correct errors. We use the base configuration described by \newcite{vaswani2017attention} but only train our model for 5,000 steps, as this problem is simpler than machine translation. We use separate vocabularies for source and target, and for efficiency consider only words in the corpus with multiple occurrences (others as $<$UNK$>$). 
Hyperparameters for this model are given in Table \ref{transformer-hyperparam-table}. 
As an evaluation metric, we rely on BLEU scores \cite{papineni2002bleu} using the correct sentence as reference. We also report the BLEU scores for the original noisy data (i.e., a baseline that does not make any correction) to show the improvement. 
The results are provided in Table \ref{tab:correction-eval}.
We observe substantial improvements on the Amazon dataset. 
However, the performance suffers on IMDB owing to its longer sequence lengths and limited training data. This suggests that our novel datasets will be useful in encouraging further research on this task.

\section{Qualitative Analysis of Generated Data}

Table \ref{error-level-table} compares the different levels of introduced errors. With a low error rate, the sentence is still intelligible with a few spelling errors. However, the quality quickly deteriorates as the error rate increases. At 15\% error rate by character, the sentence becomes very hard to understand even for a human. One can observe an increase in the length of the sentence as the error increases, as the rate of insertion errors is higher than for deletion errors in the statistics computed from the real-world data.

\begin{table}[htbp!]
\centering
\footnotesize
\begin{tabularx}{\columnwidth}{lX}
    \toprule
    Type &  Sentence\\
    \midrule
    Original & These coffee k cups have an exceptionally bold flavor. The value is great. We bought a second box and will continue to enjoy more in the future. \\
    Low Error Rate (3.75\%) & These coffee k cups hate an exceptionallly bold flavor. The value is great. We bought a second box and will kontinue to enjoy more in thye futurt. \\
    Medium Error Rate (7.5\%) & Thsse coffee k cups ahbe an exceeptionall bold falvor. Thye value is great. We bought a second box and wgll contikre ot enjos more in the future. \\
    High Error rate (15\%) & Thkese cvfffe uk dups ave an etxcepiionallyy bolg fladvor. The value ics great. We beught a second box and whll continnue yo renjy mre un the futere. \\
    \bottomrule
\end{tabularx}
\caption{Levels of Introduced Error}
\label{error-level-table}
\end{table}

\begin{table}[htbp!]
\centering
\footnotesize
\begin{tabularx}{\columnwidth}{XX}
    \toprule
    Type &  Sentence\\
    \midrule
    Original & These coffee k cups have an exceptionally bold flavor. The value is great. We bought a second box and will continue to enjoy more in the future. \\
    Medium Error Rate (7.5\%) & Thsse coffee k cups ahbe an exceeptionall bold falvor. Thye value is great. We bought a second box and wgll contikre ot enjos more in the future. \\
    Enchant Spell Corrector (without confusion enforced) & These coffee k cups ah an exceptional bold flavor. Tye value is great. We bought a second box and well continent OT enjoys more in the future. \\
    Enchant Spell Corrector (with confusion enforced) & Those coffee k cups ah an exceptional bold flavor. Tye value is great. We bought a second box and well continent OT enjoys more in the future. \\
    \bottomrule
\end{tabularx}
\caption{Result from Dictionary-based Spelling Correction}
\label{error-correction-table}
\end{table}

\begin{table}[htbp!]
\centering
\footnotesize
\begin{tabularx}{\columnwidth}{XX}
\toprule
Type &  Sentence\\
\midrule
Original & my baby eats these like they are going out of style they are the perfect size for her to grasp with her fingers so nutritional too   \\
Medium Error Rate (7.5\%)  & my baby eats these likx thre are going oht of style tehy ared the perfect sjze for her yo grasp wivh herv fingers so nutriitional to \\
Enchant Spell Corrector (with confusion enforced) & my baby eats these alix thee are going hot of style thy ares the perfect saxe for her yew grasp wive herb fingers so malnutrition to \\
Corrected with Transformer (from Enchant correction) & my baby eats these like they are going hot of style they are the perfect size for her to grasp with her fingers so nutritional too  \\
\midrule
Original & don't the sellers read these reviews and say something to the manufacturer it is a terrible that amazon can sell this product to the public \\
Medium Error Rate (7.5\%) & do't the sellers read these revies nad say something to he manufacturer pit is a rerrible that amazon can ysle ths prxduct to hte public \\
Enchant Spell Corrector (with confusion enforced) & dot the sellers read these revise bad say something to eh manufacturer pit is a revertible that amazon can isle th duct to ht public  \\
Corrected with Transformer (from Enchant correction) & got the sellers read these reviews and say something to the manufacturer it is a terrible that amazon can ship the product to the public  \\
\midrule
Original & i bought this for my niece for christmas she loves it the bamboo showed up intact and looking gorgeous and green in the middle of december   \\
Medium Error Rate (7.5\%) & i booqhtt htis for my niece for chrustmas she loves it the bmbou showed up intact and looking gorgeous antd creen in the wmiddle fo dacember \\
Enchant Spell Corrector (with confusion enforced) & i booth hits for my niece for christmas she loves it the bomb showed up intact and looking gorgeous ants screen in the middle few december   \\
Corrected with Transformer (from Enchant correction) & i bought this for my niece for christmas she loves it the combo showed up intact and looking gorgeous and screen in the middle of december \\
\bottomrule
\end{tabularx}
\caption{Randomly Selected Examples of Context-Aware Spelling Correction}
\label{error-correction-table-transformer}
\end{table}

Table \ref{error-correction-table} shows the output of the Enchant spell corrector on a sentence corrupted with medium error rate. For the given sentence, we get the same result for both with and without enforcing confusion. This happens when Enchant does not suggest the correct word for any of the errors. These errors can manifest themselves as out-of-context words such as ‘well’ and ‘continent’ in the given example or as a grammatical mistake, where ‘enjoy’ becomes ‘enjoys’.

Table \ref{error-correction-table-transformer} provides a few examples of text corruption and subsequent corrections for qualitative comparison. The samples were picked randomly from the medium error variant of the Amazon dataset. \footnote{A constraint was placed to not select sentences longer than 30 words due to space limitations.} 
We can see that the error correction model is quite accurate at simple corrections like `wive herb fingers' to `with her fingers' in example 1, and `i booth hits' to `i bought this' in example 2. 
It is also able to accurately predict complex correction requiring more context, e.g., `malnutrition' is correctly predicted to be `nutritional' given the positive sentiment of the review. 
In a more ambiguous situation where the exact word is harder to predict, the model still makes a sensible prediction. For instance, `bomb' is corrected to `combo' in example 3, and `isle' to the verb `ship' in example 2, which, albeit not correct, is a sensible prediction in the context, similar in meaning to the ground truth word `send'.. 
However, it often fails to correct obvious errors, especially when the out-of-context word is a fairly common one. For example, `hot of style' in example 1 and `gorgeous and screen' in example 3 are left uncorrected.
This model is entirely independent of the error detection model. A two-component design feeding the prediction of the detection model into the correction model may improve the results.

\section{Conclusion}

This paper proposes a method to induce typographical errors based on realistic error modeling, which is used to induce two novel typographic error datasets from different domains, each generated at three different error levels.
We show that BiLSTM neural networks are fairly effective in detecting these errors and that Transformer networks
show potential in correcting them.
Our data is freely available online at \url{http://typo.nlproc.org} for the community to make further progress on this challenging task.

\section{References}
\label{main:ref}
\bibliographystyle{lrec}
\bibliography{references}

\end{document}